\documentclass[lettersize,journal]{IEEEtran}
\usepackage{amsmath,amsfonts}
\usepackage{algorithmic}
\usepackage{algorithm}
\usepackage{array}
\usepackage[caption=false,font=footnotesize,labelfont=rm,textfont=rm]{subfig}
\usepackage{textcomp}
\usepackage{stfloats}
\usepackage{url}
\usepackage{verbatim}
\usepackage{graphicx}
\usepackage{cite}
\usepackage{minted}
\usepackage{multirow}
\usepackage{multicol}
\usepackage[table]{xcolor} 
\usepackage{colortbl}
\usepackage{makecell}
\usepackage{booktabs}
\usepackage{threeparttable}
\usepackage[colorlinks,
            linkcolor=red,
            anchorcolor=blue,
            citecolor=green
            ]{hyperref}

\begin{document}

\title{Evolution-Inspired Sample Competition for Deep Neural \\Network Optimization}

\author{Ying Zheng, Yiyi Zhang, Yi Wang,~\IEEEmembership{Member,~IEEE}, and Lap-Pui Chau$^\ast$,~\IEEEmembership{Fellow,~IEEE}% <-this % stops a space
\thanks{The research work was conducted in the JC STEM Lab of Machine Learning and Computer Vision funded by The Hong Kong Jockey Club Charities Trust. This research received partially support from the Global STEM Professorship Scheme from the Hong Kong Special Administrative Region. This work was supported in part by the National Natural Science Foundation of China (No. 62106236).}% <-this % stops a space
\thanks{Ying Zheng, Yi Wang and Lap-Pui Chau are with the Department of Electrical and Electronic Engineering, The Hong Kong Polytechnic University, Hong Kong, China. E-mail: \{ying1.zheng, yi-eie.wang, lap-pui.chau\}@polyu.edu.hk.}% <-this % stops a space
\thanks{Yiyi Zhang is with the Department of Computer Science and Engineering, The Chinese University of Hong Kong, Hong Kong, China. E-mail: yyzhang24@cse.cuhk.edu.hk.}% <-this % stops a space
\thanks{$\ast$ denotes corresponding author.}}

% The paper headers
\markboth{Evolution-Inspired Sample Competition for Deep Neural Network Optimization}%
{Shell \MakeLowercase{\textit{et al.}}: A Sample Article Using IEEEtran.cls for IEEE Journals}

\maketitle

\begin{abstract}
Conventional deep network training generally optimizes all samples under a largely uniform learning paradigm, without explicitly modeling the heterogeneous competition among them. Such an oversimplified treatment can lead to several well-known issues, including bias under class imbalance, insufficient learning of hard samples, and the erroneous reinforcement of noisy samples. In this work, we present \textit{Natural Selection} (NS), a novel evolution-inspired optimization method that explicitly incorporates competitive interactions into deep network training. Unlike conventional sample reweighting strategies that rely mainly on predefined heuristics or static criteria, NS estimates the competitive status of each sample in a group-wise context and uses it to adaptively regulate its training contribution. Specifically, NS first assembles multiple samples into a composite image and rescales it to the original input size for model inference. Based on the resulting predictions, a natural selection score is computed for each sample to characterize its relative competitive variation within the constructed group. These scores are then used to dynamically reweight the sample-wise loss, thereby introducing an explicit competition-driven mechanism into the optimization process. In this way, NS provides a simple yet effective means of moving beyond uniform sample treatment and enables more adaptive and balanced model optimization. Extensive experiments on 12 public datasets across four image classification tasks demonstrate the effectiveness of the proposed method. Moreover, NS is compatible with diverse network architectures and does not depend on task-specific assumptions, indicating its strong generality and practical potential. The code will be made publicly available.

\end{abstract}


\section{Introduction}
\label{sec:introduction}
\IEEEPARstart{N}{atural} ecosystems are characterized by the coexistence of multiple species, genetic diversity within populations, and persistent competition for limited resources. According to Darwin's theory of natural selection \cite{darwin1859origin}, individuals with greater fitness are more likely to survive, reproduce, and transmit their genes, thereby driving species toward higher levels of adaptation and fitness. Throughout this evolutionary process, both interspecific and intraspecific competition continuously influence the ecosystem's dynamic patterns.

Analogously, the optimization process of deep networks can be regarded as a mechanism of selection, evolution, and adaptation. Given a dataset with multiple categories, the network updates its parameters continuously during iterative optimization. Each forward and backward pass acts as a \textit{selection} step, where the system computes the loss for each sample and determines parameter updates that minimize the overall training error based on gradients. These updates represent a form of \textit{evolution}, as the network progressively adjusts its parameters to improve performance.  Over many iterations, the parameters gradually converge toward an optimal solution that accommodates the majority of samples. This process is similar to how biological populations \textit{adapt} to their environment.

However, classical training methods, such as minimizing the cross-entropy loss with stochastic gradient descent (SGD) \cite{lecun2002gradient,goodfellow2016deep}, essentially impose a uniform selective pressure on all ``species (categories)'' and ``individuals (samples)'' without distinction, failing to adequately account for the differences in competition among samples. In other words, although mainstream network optimization strategies are powerful, they may serve as an inefficient substitute for the complex real evolutionary processes we aim to simulate. This simplification further leads to various issues in network training, such as preference bias caused by class imbalance \cite{pei2023survey}, inadequate learning of hard samples \cite{lin2017focal}, and the misselection of noisy samples \cite{song2022learning}. Although strategies such as class reweighting \cite{cui2019class} and hard example mining \cite{shrivastava2016training} can alleviate imbalance and hard-sample issues, their heuristic nature demands careful tuning, introduces prior sensitivity, and risks amplifying noise.

To better incorporate sample heterogeneity into model optimization, we propose an evolution-inspired training method termed \emph{Natural Selection} (NS). The key idea is to explicitly model sample competition and use the resulting competitive signals to regulate training. Specifically, a group of samples is stitched and scaled before being fed into the network for prediction. Based on the prediction responses, we compute a natural selection score for each sample, which measures its competitive variation under the group context. This score is then used to dynamically adjust the sample-wise training loss, allowing the model to assign training emphasis in a more adaptive and balanced manner. By introducing an explicit competition mechanism into the optimization process, NS offers a simple yet effective way to move beyond uniform sample treatment in standard deep learning training.

We evaluate NS on four representative computer vision tasks over 12 public datasets. Extensive experimental results demonstrate the effectiveness of the proposed method across diverse settings. Moreover, because NS operates at the training-mechanism level rather than relying on a specific architecture design, it can be naturally integrated with different backbone families, indicating its potential for broader applicability.

Our main contributions are summarized as follows:
\begin{itemize}
    \item We introduce an evolution-inspired perspective for understanding deep network optimization, emphasizing the role of sample competition in model evolution.
    \item We propose NS, a novel training method that dynamically adjusts sample-wise loss weights based on natural selection scores derived from explicit sample competition.
    \item Extensive experiments on 12 public datasets spanning four common computer vision tasks verify the effectiveness and generality of the proposed method.
\end{itemize}

\section{Related Work}
This section provides a brief overview of relevant work in data weighting, mix-based data augmentation, and evolutionary computation.

\subsection{Data weighting}
Data weighting methods \cite{wu2025data} can be broadly categorized into two types: class-level weighting and instance-level weighting. Class-level weighting assigns the same weight to all samples within a class, commonly used to alleviate class imbalance issues. Examples include the soft-weighted loss \cite{zheng2020deep} based on class frequency and the class-balanced loss \cite{cui2019class} derived from the effective number of samples, both of which aim to enhance the model's ability to recognize minority classes. In contrast, instance-level weighting \cite{kumar2010self,chang2017active} independently computes or learns a weight for each sample, allowing for a more fine-grained reflection of each sample’s importance or learning difficulty during training \cite{xia2025instance,zhou2022region}.

In recent years, instance-level weighting has emerged as the mainstream in data weighting, due to its ability to assign individualized weights to each sample, which shows effectiveness in addressing challenges such as class imbalance \cite{fernando2021dynamically,guo2022learning}, noisy labels \cite{zhang2021learning,yao2024cosw}, and domain shift \cite{gu2021adversarial,nguyen2023evolutionary}. These methods generally fall into several categories: 1) importance-based weighting \cite{kimura2024short,holstegeoptimizing}, which assigns weights by measuring the rarity of a sample within a distribution or its degree of mismatch with the target distribution; 2) meta-learning-based weighting \cite{shu2019meta,shu2023cmw}, which formulates weight learning within a bi-level optimization framework, guiding the training of the weighting network by optimizing a meta-loss on a small and clean validation set; 3) difficulty-driven weighting \cite{lin2017focal,zhou2023samples}, such as focal loss \cite{lin2017focal}, which introduces a modulating factor into the cross-entropy loss to suppress gradients from easy samples and amplify learning signals from hard ones. It is worth noting that while most existing methods focus on weighting original training samples, some studies have also explored assigning weights to augmented samples \cite{yireweighting,han2022umix}. Distinct from the aforementioned approaches, this paper proposes a novel instance-level weighting method inspired by natural selection, which dynamically adjusts the weights of dominant and disadvantaged samples during competition, thereby achieving more effective model training.

\subsection{Mix-based data augmentation}
In this work, we utilize image stitching to create synthetic images, a technique that shares some similarities with mix-based data augmentation methods \cite{cao2024survey}. As a pioneering data augmentation technique, mixup \cite{zhang2018mixup} enhances model generalization by interpolating both images and labels of different samples. Subsequently, researchers have proposed a variety of mixing strategies, such as CutMix \cite{yun2019cutmix}, RICAP \cite{takahashi2019data}, RankMixup \cite{noh2023rankmixup}, and Similarity Kernel Mixup \cite{bouniottailoring}, which have demonstrated promising effectiveness in improving model robustness and generalization. Although our method also relies on image synthesis, its motivation and objective are fundamentally different from those of the aforementioned data augmentation approaches. Rather than aiming to increase data diversity, we leverage synthetic images to simulate the mechanism of natural selection, with the goal of adjusting the learning weights assigned to samples with varying competitive advantages, thereby more effectively guiding the model to focus on differences among samples during training.

\subsection{Evolutionary Computation}
Evolutionary computation has become a crucial evolution-inspired paradigm in machine learning and optimization research, owing to its flexibility, population-based search mechanism, and strong exploratory capabilities \cite{zhang2011evolutionary,bi2022survey}. These characteristics make evolutionary methods particularly suitable for complex optimization scenarios that require effective search and adaptation.

More recently, increasing attention has been devoted to integrating evolutionary algorithms with reinforcement learning, resulting in hybrid frameworks that leverage the complementary strengths of both paradigms in policy search, exploration, and optimization \cite{li2024bridging}. Existing studies have investigated such combinations in various settings, including reinforcement learning with reputation-based interaction \cite{ren2023reputation}, action learning for deep reinforcement learning \cite{xu2024niching}, and reinforcement-learning-inspired simulation optimization \cite{preil2024genetic}. In parallel, evolutionary methods have also demonstrated a promising capability in neural architecture search, enabling the automatic discovery of effective deep neural network structures \cite{zhang2022evolutionary,cao2025comprehensive}. Inspired by these developments, our method incorporates biologically motivated principles to explore a new evolutionary approach to deep neural network optimization.

\section{Methodology}
In this section, we first reinterpret deep neural network optimization through an evolutionary lens, then introduce natural selection over training samples, and finally present the corresponding optimization method.

\subsection{Rethinking Neural Network Optimization through an Evolutionary Lens}
The optimization process of deep neural networks exhibits a profound yet often overlooked similarity with the natural evolution of species in ecosystems. For instance, in supervised learning for image classification, the training process of a neural network can be analogized to an evolutionary process within a ``digital ecosystem''. To formalize this analogy, we first introduce the supervised learning setting as follows:
\begin{itemize}
    \item Let $\mathcal{X}$ be the input space and $\mathcal{C} = \{1, \ldots, K\}$ represent the set of $K$ discrete class labels. Given a training dataset $\mathcal{D} = \{(x_i, y_i)\}_{i=1}^N$, each $x_i \in \mathcal{X}$ represents an input instance and is paired with a corresponding class label $y_i \in \mathcal{C}$, and $N$ denotes the total number of samples.
    \item A classifier parameterized by $\theta$ is defined as a function $f_\theta: \mathcal{X} \to \mathbb{R}^K$, which outputs a score or probability vector over the $K$ classes for each input. For a given training sample $(x_i, y_i)$, the loss is computed as $\ell_i(\theta) = \ell(f_\theta(x_i), y_i)$, where $\ell(\cdot, \cdot)$ denotes a chosen loss function, such as the cross-entropy loss.
    \item The overall training objective is to minimize the empirical risk: $L(\theta) = \frac{1}{N} \sum_{i=1}^N \ell_i(\theta)$. This objective is typically optimized using iterative optimization algorithms, such as stochastic gradient descent (SGD) or Adam.
\end{itemize}

To establish a mathematical foundation for the evolutionary analogy, we define a mapping between the components of supervised learning and evolutionary concepts:
\begin{itemize}
    \item \textbf{Population structure}: The training dataset $\mathcal{D}$ corresponds to a population of individuals, where each sample $(x_i, y_i)$ represents an individual.
    \item \textbf{Fitness function}: We define the fitness of an individual as $h_i(\theta) = M - \ell_i(\theta)$, where $M$ is a constant ensuring $f_i(\theta) > 0$. This suggests that high fitness is associated with low loss.
    \item \textbf{Selective pressure}: The gradient $\nabla_\theta \ell_i(\theta)$ provides directional pressure analogous to natural selection, driving the system toward fitter states.
\end{itemize}

\textbf{Proposition 1 (Evolutionary-Learning Correspondence).} Under the mapping $\phi$ defined by:
\begin{align*}
    \phi(\text{population}) &= \mathcal{D}, \\
    \phi(\text{individual } i) &= (x_i, y_i), \\
    \phi(\text{fitness of individual } i) &= M - \ell_i(\theta), \\
    \phi(\text{selective pressure}) &= \nabla_\theta \ell_i(\theta),
\end{align*}
there exists a structural correspondence between the evolutionary process maximizing population fitness and the learning process minimizing empirical risk, characterized by equivalent optimization objectives and similar iterative improvement dynamics.

We establish the structural correspondence through two key aspects:

1) \textbf{Objective function duality}. The evolutionary objective of maximizing average fitness:
\begin{equation}
   \max_{\theta} \frac{1}{N} \sum_{i=1}^N h_i(\theta) = \max_{\theta} \frac{1}{N} \sum_{i=1}^N [M - \ell_i(\theta)]
\label{eq:maxfi}
\end{equation}
is mathematically equivalent to the learning objective of minimizing empirical risk:
\begin{equation}
   \min_{\theta} \frac{1}{N} \sum_{i=1}^N \ell_i(\theta)
\label{eq:minli}
\end{equation}
since maximizing $M - \ell_i(\theta)$ is equivalent to minimizing $\ell_i(\theta)$ for constant $M$.

2) \textbf{Improvement dynamics}. Both processes employ iterative improvement strategies: In evolution, fitter individuals are more likely to reproduce, gradually improving population fitness; In learning, parameters are updated in the direction that reduces loss: $\theta_{t+1} = \theta_t - \eta \cdot \nabla_\theta L(\theta)$. While their update process differ, both mechanisms drive the system toward better performance over time.

This formal correspondence provides a mathematical foundation for interpreting network optimization through an evolutionary lens, highlighting the structural similarities between biological evolution and learning optimization.

By establishing an analogy between network optimization and ecosystem evolution, we can critically re-examine the limitations of the classical supervised learning paradigm with Empirical Risk Minimization (ERM). The standard ERM framework minimizes the average loss over the training dataset: $\min_\theta \frac{1}{N}\sum_{i=1}^N \ell(x_i, y_i; \theta)$, which implicitly assigns equal weight $w_i = 1$ to every sample $(x_i, y_i)$. This uniform weighting scheme, while statistically well-motivated, fails to account for the competitive dynamics among samples. In ecological terms, the ERM objective applies a constant selective pressure $\nabla_\theta \ell(x_i, y_i; \theta)$ to all individuals, regardless of their fitness or the population composition. The gradient signal that drives optimization, $\frac{1}{N}\sum_{i=1}^N \nabla_\theta \ell(x_i, y_i; \theta)$, represents a global average that lacks the fine-grained adaptation to individual competitive advantages.

The core limitation lies in the homogeneity of the weighting function $w_i \equiv 1$, which prevents the emergence of dynamic selection pressures analogous to those in natural ecosystems. Consequently, the network optimization process cannot prioritize samples with higher fitness or protect developing minority patterns $\mathcal{D}_{\text{min}}$ from being overwhelmed by dominant patterns $\mathcal{D}_{\text{maj}}$. From an evolutionary perspective, this eliminates the natural selection mechanism essential for maintaining diversity and adapting to complex environments, often leading to premature convergence and limited generalization capability.

In essence, by neglecting individual-level competition mechanisms akin to natural selection, classical training methods are thus more prone to issues such as local optima, overfitting, or underfitting. \textit{In paradigms lacking such competition, can Darwinian natural selection be incorporated into the training process to restructure the optimization dynamics?}

\begin{figure*}[t]
    \centering
    {\includegraphics[width=0.9\linewidth]{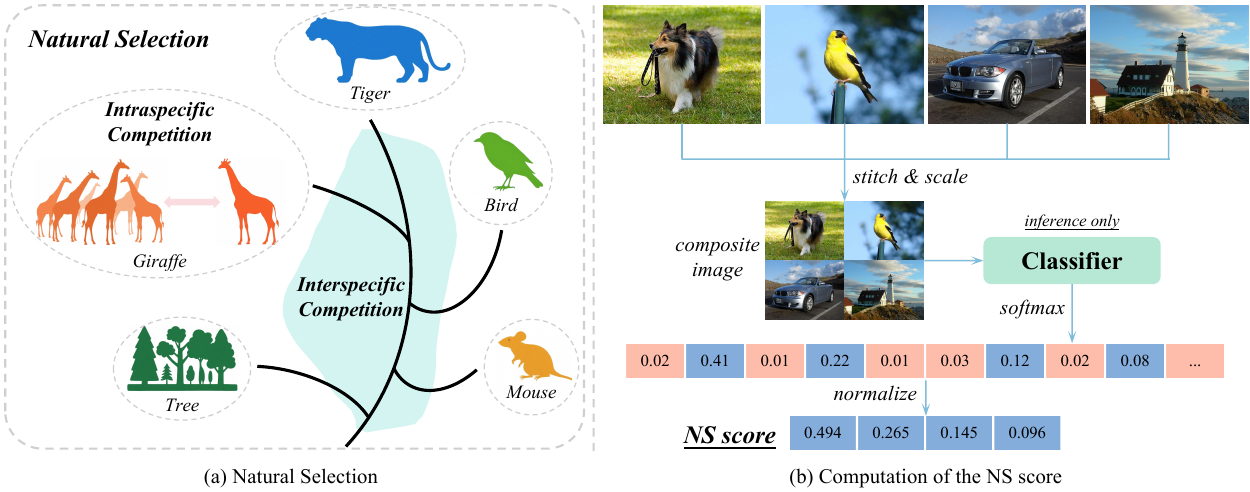}}
    \caption{(a) Natural selection in ecosystem evolution. (b) Illustration of the natural selection process in training samples. After performing image stitching and scale normalization, the resulting stitched image is fed into the classifier to obtain the softmax probability distribution. The predicted probability associated with each sample’s category is then extracted, and the NS score is subsequently computed through normalization.}
    \label{fig:illustration}
\end{figure*}

\subsection{Natural Selection of Training Samples}
Before addressing the above question, we first revisit the mechanism of natural selection in nature. As shown in Fig.~\ref{fig:illustration} (a), natural selection is primarily driven by two-fold competition: 1) interspecific competition, in which species compete for limited resources, and 2) intraspecific competition, where individuals within a species compete for survival and reproduction. The core of both types of competition lies in the competition among individuals. Inspired by this mechanism, we introduce similar competitive relations into network training: samples compete for update resources according to their current competitive status and are subjected to varying degrees of selective pressure, thereby constructing an ecological dynamic balance during training.

To simulate individual competition and quantify the relative advantage of each sample, we propose a \emph{Natural Selection} (NS) mechanism that assigns each sample an NS score through explicit intra-group competition. As illustrated in Fig.~\ref{fig:illustration}(b), given a group of training samples $\mathcal{G}=\{(x_i,y_i)\}_{i=1}^{m}$ (e.g., $m=4$), where each sample $x_i \in \mathbb{R}^{H_0 \times W_0 \times 3}$ and $y_i \in \{1,\dots,K\}$, we first arrange the $m$ samples into an $R \times C$ grid with $R \times C = m$, and concatenate them along the spatial dimensions to construct a composite image:
\begin{equation}
    S = \text{Stitch}(x_1, x_2, \dots, x_m) \in \mathbb{R}^{R \cdot H_0 \times C \cdot W_0 \times 3}.
\label{eq:stitch}
\end{equation}
The composite image is then resized to the original input resolution via bilinear interpolation, yielding $S' \in \mathbb{R}^{H_0 \times W_0 \times 3}$. The resized sample $S'$ is fed into the classifier $f_{\theta}(\cdot)$ to produce the logits:
\begin{equation}
z = f_{\theta}(S') \in \mathbb{R}^{K},
\end{equation}
followed by the posterior probability vector:
\begin{equation}
p=\mathrm{softmax}(z)\in[0,1]^K, \qquad \sum_{k=1}^{K} p_k = 1.
\end{equation}
Here, $f_{\theta}(\cdot)$ is used only for forward inference and is detached from the standard optimization pipeline. Therefore, the NS-score computation introduces no additional gradient flow to the main training process.

For each sample $x_i \in \mathcal{G}$, we use the posterior probability of its ground-truth class as its raw competitive score:
\begin{equation}
q_i = p[y_i].
\label{eq:raw_score}
\end{equation}
To induce explicit competition within the group and suppress scale variation across different groups, we further normalize the raw scores and define the NS score as:
\begin{equation}
s_i = \frac{q_i}{\sum_{j=1}^{m} q_j}.
\label{eq:ns_score}
\end{equation}
A larger $s_i$ indicates that sample $x_i$ exhibits a stronger competitive advantage within the group (the winner), whereas a smaller $s_i$ corresponds to a weaker one (the loser). In this way, the score vector $\mathbf{s} = (s_1,\dots,s_m)$ characterizes the relative competitiveness among samples and serves as the basis for imposing selective pressure during training. The overall computation of NS scores is summarized in Algorithm~\ref{alg:natural_selection}.

\begin{algorithm}[t]
\caption{Natural Selection Score Computation}
\label{alg:natural_selection}
\textbf{Input}: Mini-batch $\{(x_b,y_b)\}_{b=1}^{B}$, classifier $f_{\theta}$, group size $m$ \\
\textbf{Output}: NS scores $\mathbf{s}\in\mathbb{R}^{B}$
\begin{algorithmic}[1]
\STATE Partition the mini-batch into $G=B/m$ groups.
\FOR{$g=1$ to $G$}
    \STATE Construct the composite image $S^{(g)}$ from $\mathcal{G}^{(g)}$ by Eq.~(\ref{eq:stitch}), and resize it to $S'^{(g)}$.
    \STATE Compute the posterior vector
    \[
    p^{(g)}=\mathrm{softmax}\!\left(f_{\theta}(S'^{(g)})\right).
    \]
    \STATE For each $(x_i,y_i)\in\mathcal{G}^{(g)}$, compute $q_i^{(g)}$ by Eq.~(\ref{eq:raw_score}) and $s_i^{(g)}$ by Eq.~(\ref{eq:ns_score}).
\ENDFOR
\STATE Collect all $\{s_i^{(g)}\}$ as $\mathbf{s}$ and return $\mathbf{s}$.
\end{algorithmic}
\end{algorithm}

\subsection{Optimizing Deep Networks via Natural Selection}
After obtaining the NS score, how to leverage it to optimize deep networks becomes a critical issue. An intuitive approach is to use the NS score as a sample weight to modulate its loss, thereby introducing differentiated selective pressure. However, this leads to a fundamental dilemma regarding how the weighting should be allocated: \textit{Should we strengthen the winners or focus more on the losers?} Prioritizing the winners would mean optimizing those that have already demonstrated stronger competitive advantages, whereas emphasizing the losers would imply assigning greater importance to samples that are in an unfavorable position during competition throughout optimization.

To address this issue, we propose the following two weighting strategies based on the NS score, each tailored to different learning scenarios:

1) \textbf{NS-based Winner-Strengthening} (NS-WS) strategy assigns larger weights to samples with higher NS scores, aiming to consolidate samples that already exhibit stronger competitive advantages. This strategy enhances the stability of gradient signals, promotes the sharpening of decision boundaries, and facilitates the convergence of optimization. Specifically, when the NS score effectively reflects sample learnability or label reliability, emphasizing high-score samples helps reduce disturbances caused by low-quality samples, making training more stable and efficient. NS-WS is particularly suitable for scenarios requiring high discriminability and stable decision boundaries. In tasks involving label noise or annotation ambiguity, it can still be effective provided that a positive correlation exists between NS scores and sample reliability, as it suppresses the adverse impact of potentially mislabeled samples on the optimization trajectory.

2) \textbf{NS-based Loser-Focusing} (NS-LF) strategy assigns greater weights to samples with lower NS scores to enhance their influence on parameter updates. Intuitively, samples with low NS scores are disadvantaged in the current competition, which may correspond to instances from long-tail categories, boundary cases, or examples that have not yet been sufficiently learned. By increasing the relative weight of such samples, the model allocates more capacity during training to rectify these under-optimized areas, thereby improving performance on minority classes and challenging regions, and enhancing overall generalization and robustness. This strategy is suitable for scenarios that require fairness or long-tail adaptation, such as tasks with imbalanced class distributions, where boundary samples significantly affect performance, or during early training phases where broader exploration is desired to avoid over-reliance on easy examples. Importantly, the effectiveness of NS-LF depends on the assumption that the NS score reflects samples that are learnable yet currently suppressed rather than those dominated by noise or annotation errors.

It should be noted that both strategies entail potential risks. NS-WS may overly rely on samples with high NS scores, resulting in selection bias that overlooks disadvantaged samples and minority classes, thereby limiting generalization. Conversely, NS-LF could amplify the influence of noisy and mislabeled samples, leading to training instability. To mitigate these issues, we introduce a base value $\sigma$ to adjust the numerical range of NS mapping and provide a lower bound for the weight of each sample. Formally, the final per-sample weight is defined as:
\begin{equation}
    w_i = \sigma + \rho \cdot s_i,
\label{eq:weight}
\end{equation}
where $\rho \in \{+1, -1\}$ denotes the strategy polarity: Setting $\rho = +1$ yields the NS-WS strategy, while choosing $\rho = -1$ implements to the NS-LF strategy. 

\begin{figure}[t]
    \centering
    \includegraphics[width=1.0\linewidth]{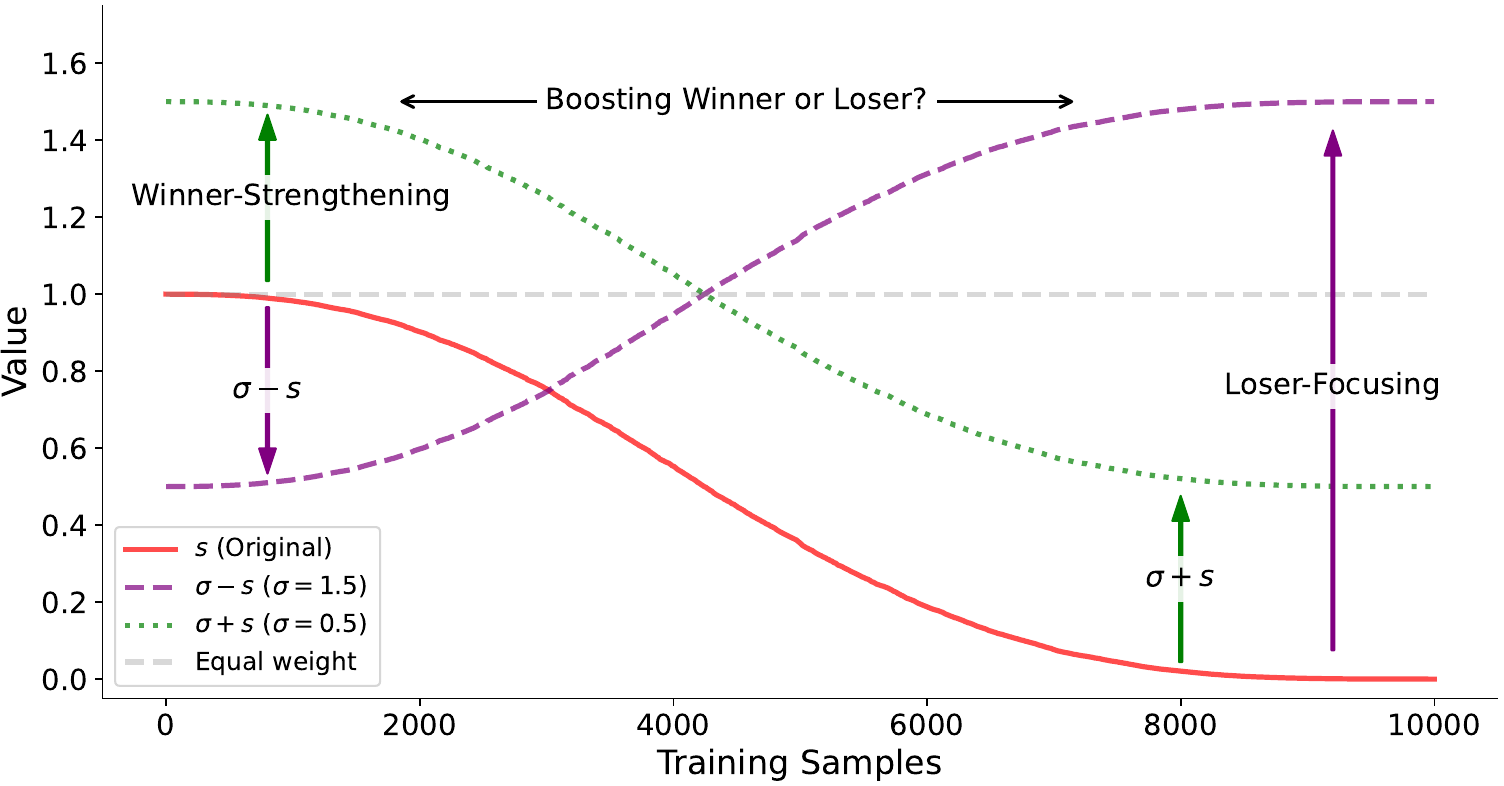}
    \caption{The distribution curves of sample weights under different settings. Under the NS-WS (green dashed line) and NS-LF (purple dashed line) strategies, samples with higher-weight values are reinforced during training, while those in lower-weight regions are suppressed. This enables differentiated optimization for samples with varying competitive characteristics.}
    \label{fig:boosting}
\end{figure}

As illustrated in Fig. \ref{fig:boosting}, we compare the weight distribution curves simulated over 10,000 samples under three settings: 1) The red solid curve represents the original mapping in descending order. In this case, the weight of samples with low NS scores $s_i$ approaches zero, resulting in negligible gradient contributions during training and thus limiting the model’s ability to learn from these examples. 2) The green dashed curve corresponds to the NS-WS strategy ($\rho = +1$) with an additive base value $\sigma$. This term preserves the relative ordering among samples while effectively smoothing the weight distribution, preventing individual samples from being assigned near-zero weights or allowing extremely high-score samples to dominate the update. 3) The purple dashed curve denotes the NS-LF strategy ($\rho = -1$), which globally increases the relative weight of low-score samples and suppresses the influence of high-score ones, while $\sigma$ ensures that the weights remain within a stable range and avoid extreme values, thereby promoting training stability.

After obtaining the sample weights from Eq. (\ref{eq:weight}), we apply them directly in the per-sample training loss, defining the weighted loss as $\ell_i(\theta) = w_i \cdot \ell(f_\theta(x_i), y_i)$. Since the calculation of the NS score is independent of the training process, this mechanism can be seamlessly integrated into existing training frameworks to enable competition-aware dynamic optimization of deep networks.

\section{Experiments}
We conduct comprehensive experiments to evaluate the effectiveness, generality, robustness, and efficiency of the proposed NS method. Specifically, we validate NS on 12 public benchmarks covering four representative computer vision tasks. We compare NS against a diverse set of loss-based and sampling-based baselines, and further provide ablation studies, statistical stability analysis, computational overhead evaluation, and additional results on hard-sample recognition and text classification.

\subsection{Experimental Setup}

\subsubsection{Datasets}
We evaluate the proposed method on 12 public benchmarks covering four mainstream computer vision tasks: image classification, visual emotion recognition, source-free domain adaptation, and long-tailed classification. 
\begin{itemize}
    \item For image classification, we use CIFAR-10 and CIFAR-100 \cite{krizhevsky2009learning}, both containing 60,000 images of size 32$\times$32 over 10 and 100 classes, respectively, and ImageNet-1K \cite{russakovsky2015imagenet}, which contains approximately 1.28 million training images and 50,000 validation images from 1,000 classes.
    \item For visual emotion recognition, we adopt six public datasets: Twitter I \cite{you2015robust}, Twitter II \cite{borth2013large}, Flickr \cite{katsurai2016image}, Instagram \cite{katsurai2016image}, FI \cite{you2016building}, and EmoSet \cite{yang2023emoset}. Following \cite{zheng2025fuzzy}, we use cleaned versions with duplicate and corrupted samples removed. Twitter I, Twitter II, Flickr, and Instagram are binary sentiment datasets with 1,251, 545, 42,848, and 60,729 images, respectively. FI and EmoSet are eight-class emotion datasets with 21,829 and 118,102 images, respectively.
    \item For source-free domain adaptation, we use Office-Home \cite{venkateswara2017deep}, which comprises 15,500 images from 65 categories across four domains: Artistic (Ar), Clipart (Cl), Product (Pr), and Real-World (Rw).
    \item For long-tailed classification, we adopt CIFAR-LT-10 and CIFAR-LT-100 \cite{cui2019class}, which are derived from CIFAR-10 and CIFAR-100 by imposing exponential class imbalance on the training set, while keeping the original balanced test set unchanged.
\end{itemize}

\subsubsection{Implementation Details}
Unless noted otherwise, all experiments are implemented in PyTorch and conducted on NVIDIA RTX 4090 GPUs. For fair comparison with prior work, we adopt widely used backbones across different tasks, including CNNs (AlexNet \cite{krizhevsky2012imagenet}, VGG \cite{simonyan2015very}, ResNet \cite{he2016deep}, ResNeXt \cite{xie2017aggregated}, and DenseNet \cite{huang2017densely}), Transformers (ViT \cite{dosovitskiy2020image} and Swin Transformer \cite{liu2021swin}), and the state-space architecture VMamba \cite{liu2024vmamba}. All reported results are averaged over three independent runs.

For image classification, we use AlexNet, VGG-19, ResNet-110, and ResNeXt-29 on CIFAR-10/100. All models are optimized using SGD with momentum of 0.9, a batch size of 128, and an initial learning rate of 0.1. AlexNet, VGG-19, and ResNet-110 are trained for 164 epochs, with the learning rate decayed by a factor of 0.1 at epochs 81 and 122. ResNeXt-29 is trained for 300 epochs, with the learning rate decayed by a factor of 0.1 at epochs 150 and 225. On the ImageNet-1K dataset, we choose ResNet-18, Swin-T, and VMamba-T as representative baseline networks, and train all models for 90 epochs. ResNet-18 is optimized by SGD with momentum 0.9, batch size 256, and initial learning rate 0.1, with decay by a factor of 0.1 at epochs 31 and 61. Swin-T and VMamba-T are trained on four NVIDIA RTX 4090 GPUs using AdamW \cite{kingma2015adam} with weight decay 0.05, cosine learning rate decay, five epochs of linear warmup, global batch size 1024, and initial learning rate 0.001. For our method, we adopt NS-WS on all three datasets, compute the NS score with groups of four samples, and set $\sigma$ to 0.7, 0.8, and 2.0 for CIFAR-10, CIFAR-100, and ImageNet-1K, respectively.

For visual emotion recognition, we use AlexNet, VGG-16, ResNet-50, DenseNet-121, ViT-B/16, and Swin-T as backbone networks. All models are initialized from ImageNet-pretrained weights and fine-tuned for 10 epochs using SGD with momentum 0.9, batch size 64, initial learning rate \(10^{-3}\), and polynomial learning rate decay. Considering the class imbalance in emotion recognition datasets, we consistently adopt NS-LF. Specifically, the NS score is computed using groups of two samples and $\sigma=2.5$ on Twitter I and Twitter II, groups of four samples and $\sigma=2.0$ on Flickr and Instagram, and groups of four samples and $\sigma=2.5$ on FI and EmoSet.

For source-free domain adaptation, following \cite{liang2020we,zheng2026procal}, we use ResNet-50 as the feature backbone and a classifier composed of a bottleneck layer (a fully connected layer followed by batch normalization) and a weight-normalized fully connected layer. The source model is first trained in a supervised manner on the source domain, after which target adaptation is performed for 15 epochs. We use SGD with momentum 0.9 and batch size 64, and set the learning rates to \(10^{-3}\) and \(10^{-2}\) for the backbone and bottleneck, respectively. The final fully connected classification layer, serving as the source hypothesis, is frozen during adaptation. To improve robustness to pseudo-label noise, we adopt NS-WS with groups of four samples and $\sigma=0.6$.

For long-tailed classification, we use ResNet-32 on CIFAR-LT-10 and CIFAR-LT-100. Training is performed for 240 epochs using SGD with momentum 0.9 and a batch size of 128. The initial learning rate is set to 0.1 and decayed at epochs 100, 160, and 200 by factors of 0.1, 0.1, and 0.01, respectively. Given the intrinsic class imbalance of long-tailed data, we adopt NS-LF on both datasets. The NS score is computed with groups of two samples and $\sigma=2.5$ on CIFAR-LT-10, and with groups of four samples and $\sigma=1.5$ on CIFAR-LT-100.

\begin{table}[t]
\centering
\caption{Evaluation of NS on \textbf{ImageNet-1K}. ``WS'' denotes boosting the winner, while ``LF'' means enhancing the loser. ``2/4'' indicates grouping two or four samples for competition.}
\begin{tabular}{lccccc}
\toprule
\multirow{2}{*}{Method} & \multirow{2}{*}{Basic} & \multicolumn{4}{c}{w/ NS} \\
\cmidrule(lr){3-6}
{} & {} & LF-2 & WS-2 & LF-4 & WS-4 \\
\midrule
ResNet-18 \cite{he2016deep} & 68.4 & 68.4 & 68.6 & 68.4 & \textbf{68.9} (\textcolor[rgb]{0.2,0.6,0.4}{$\uparrow$0.5}) \\
Swin-T \cite{liu2021swin} & 75.2 & 75.4 & 75.5 & 75.2 & \textbf{75.7} (\textcolor[rgb]{0.2,0.6,0.4}{$\uparrow$0.5}) \\
VMamba-T \cite{liu2024vmamba} & 76.7 & 76.9 & 76.9 & 76.8 & \textbf{77.1} (\textcolor[rgb]{0.2,0.6,0.4}{$\uparrow$0.4}) \\
\bottomrule
\end{tabular}
\label{tab:imagenet}
\end{table}

\begin{table}[t]
\centering
\caption{Evaluation of NS-WS on \textbf{CIFAR-10/100} with different networks.}
\begin{tabular}{lcccc}
\toprule
\multirow{2}{*}{Method} & \multicolumn{2}{c}{\textbf{CIFAR-10}} & \multicolumn{2}{c}{\textbf{CIFAR-100}} \\
\cmidrule(lr){2-3} \cmidrule(lr){4-5}
{} & Basic & w/ NS-WS & Basic & w/ NS-WS \\
\midrule
AlexNet \cite{krizhevsky2012imagenet} & 77.4 & \textbf{78.6} (\textcolor[rgb]{0.2,0.6,0.4}{$\uparrow$1.2}) & 43.8 & \textbf{46.0} (\textcolor[rgb]{0.2,0.6,0.4}{$\uparrow$2.2}) \\
VGG-19 \cite{simonyan2015very} & 93.4 & \textbf{93.8} (\textcolor[rgb]{0.2,0.6,0.4}{$\uparrow$0.4}) & 71.7 & \textbf{72.6} (\textcolor[rgb]{0.2,0.6,0.4}{$\uparrow$0.9}) \\
ResNet-110 \cite{he2016deep} & 92.8 & \textbf{93.9} (\textcolor[rgb]{0.2,0.6,0.4}{$\uparrow$1.1}) & 70.3 & \textbf{72.1} (\textcolor[rgb]{0.2,0.6,0.4}{$\uparrow$1.8}) \\
ResNeXt-29 \cite{xie2017aggregated} & 95.9 & \textbf{96.2} (\textcolor[rgb]{0.2,0.6,0.4}{$\uparrow$0.3}) & 81.4 & \textbf{82.3} (\textcolor[rgb]{0.2,0.6,0.4}{$\uparrow$0.9}) \\
\bottomrule
\end{tabular}
\label{tab:cifar}
\end{table}

\subsubsection{Compared Methods}
We compare our method with 12 representative loss-based approaches, including Focal Loss \cite{lin2017focal}, GCE \cite{zhang2018generalized}, label smoothing (LS) \cite{muller2019does}, NLNL \cite{kim2019nlnl}, SCE \cite{wang2019symmetric}, APL \cite{ma2020normalized}, LOW \cite{santiago2021low}, PolyLoss \cite{lengpolyloss}, ANL \cite{ye2023active}, AUL \cite{zhou2023asymmetric}, CE$_\epsilon$+MAE \cite{wang2024epsilon}, and CE+OGC \cite{ye2025optimized}. These methods improve training primarily through loss redesign at the sample level. We further include three classical sampling-based baselines: Class-Balanced Sampling (CBS), which enforces class-uniform sampling; Square-Root Sampling (SRS), which moderates sampling bias using class-frequency square roots; and Progressively-Balanced Sampling (PBS), which gradually interpolates from frequency-based sampling to class-balanced sampling during training.

\subsection{Main Results}
\subsubsection{Image Classification}
We evaluate the proposed NS method on the large-scale ImageNet-1K dataset \cite{russakovsky2015imagenet} using three different types of networks. Specifically, we compare four experimental settings that combine two group sizes (2 and 4 samples per group) with two enhancement strategies (i.e., NS-WS and NS-LF). Results are summarized in Table \ref{tab:imagenet}. We empirically observe that boosting winners improves classification accuracy, whereas enhancing losers yields no significant improvement. This finding contradicts the conventional intuition that disadvantaged samples should be prioritized for enhancement. However, upon further analysis, we find this behavior reasonable for datasets like ImageNet-1K, which have relatively balanced class distributions but non-negligible label noise. Enhancing winners helps the model learn more generalizable class-specific features while mitigating the negative impact of noisy samples. In addition, results show that grouping four samples per competition achieves superior performance compared to grouping two. Overall, the experimental results demonstrate that NS-WS can consistently improve performance without relying on specific network architectures.

\begin{table*}[t]
\centering
\setlength{\tabcolsep}{1.8mm}
\caption{Comparison of basic and NS-enhanced models on six emotion recognition datasets.}
\begin{tabular}{@{}lcccccccccccc@{}}
\toprule
\multirow{2}{*}{Method} & \multicolumn{2}{c}{\textbf{Twitter I}} & \multicolumn{2}{c}{\textbf{Twitter II}} & \multicolumn{2}{c}{\textbf{Flickr}} & \multicolumn{2}{c}{\textbf{Instagram}} & \multicolumn{2}{c}{\textbf{FI}} & \multicolumn{2}{c}{\textbf{EmoSet}} \\
\cmidrule(lr){2-3} \cmidrule(lr){4-5} \cmidrule(lr){6-7} \cmidrule(lr){8-9} \cmidrule(lr){10-11} \cmidrule(lr){12-13}
{} & Basic & w/ NS-LF & Basic & w/ NS-LF & Basic & w/ NS-LF & Basic & w/ NS-LF & Basic & w/ NS-LF & Basic & w/ NS-LF \\
\midrule
AlexNet \cite{krizhevsky2012imagenet} & 73.8 & \textbf{75.2} & 68.5 & \textbf{71.0} & 83.2 & \textbf{83.8} & 80.6 & \textbf{81.2} & 56.4 & \textbf{57.9} & 67.6 & \textbf{69.2} \\
VGG-16 \cite{simonyan2015very} & 75.3 & \textbf{78.1} & 74.3 & \textbf{74.9} & 84.4 & \textbf{84.6} & 82.5 & \textbf{83.1} & 59.7 & \textbf{61.4} & 72.2 & \textbf{73.9} \\
ResNet-50 \cite{he2016deep} & 81.3 & \textbf{82.9} & 71.0 & \textbf{74.3} & 85.7 & \textbf{86.4} & 84.6 & \textbf{85.2} & 65.8 & \textbf{67.2} & 75.8 & \textbf{76.3} \\
DenseNet-121 \cite{huang2017densely} & 78.9 & \textbf{79.4} & 72.8 & \textbf{74.0} & 85.1 & \textbf{85.6} & 84.0 & \textbf{85.0} & 63.2 & \textbf{64.4} & 74.9 & \textbf{75.2} \\
ViT-B/16 \cite{dosovitskiy2020image} & 74.5 & \textbf{76.4} & 72.1 & \textbf{72.2} & 86.2 & \textbf{86.4} & 84.5 & \textbf{85.3} & 65.6 & \textbf{67.4} & 77.3 & \textbf{77.7} \\
Swin-T \cite{liu2021swin} & 69.3 & \textbf{69.7} & 70.0 & \textbf{73.1} & 83.9 & \textbf{85.9} & 82.3 & \textbf{82.5} & 64.4 & \textbf{67.0} & 76.9 & \textbf{78.4} \\
\midrule
Avg. & 75.5 & \textbf{77.0} (\textcolor[rgb]{0.2,0.6,0.4}{$\uparrow$1.5}) & 71.5 & \textbf{73.3} (\textcolor[rgb]{0.2,0.6,0.4}{$\uparrow$1.8}) & 84.8 & \textbf{85.5} (\textcolor[rgb]{0.2,0.6,0.4}{$\uparrow$0.7}) & 83.1 & \textbf{83.7} (\textcolor[rgb]{0.2,0.6,0.4}{$\uparrow$0.6}) & 62.5 & \textbf{64.2} (\textcolor[rgb]{0.2,0.6,0.4}{$\uparrow$1.7}) & 74.1 & \textbf{75.1} (\textcolor[rgb]{0.2,0.6,0.4}{$\uparrow$1.0}) \\
\bottomrule
\end{tabular}
\label{tab:emotion}
\end{table*}

\begin{table*}[t]
\centering
\setlength{\tabcolsep}{1.8mm}
\caption{Comparison of different methods on the \textbf{Office-Home} benchmark.}
\begin{tabular}{@{}lccccccccccccc@{}}
\toprule
{Method} & Ar$\rightarrow$Cl & Ar$\rightarrow$Pr & Ar$\rightarrow$Rw & Cl$\rightarrow$Ar & Cl$\rightarrow$Pr & Cl$\rightarrow$Rw & Pr$\rightarrow$Ar & Pr$\rightarrow$Cl & Pr$\rightarrow$Rw & Rw$\rightarrow$Ar & Rw$\rightarrow$Cl & Rw$\rightarrow$Pr & Avg. \\
\midrule
Baseline & 58.0 & 78.8 & 81.9 & 69.4 & 79.7 & 78.7 & 68.2 & 55.0 & 81.9 & 73.8 & 58.7 & 83.8 & 72.3 \\
Focal Loss \cite{lin2017focal} & 57.9 & 78.8 & 82.0 & 69.2 & 79.8 & 78.6 & 68.3 & 55.4 & 81.8 & 73.7 & 59.9 & 83.7 & 72.4 \\
GCE \cite{zhang2018generalized} & 57.5 & 77.7 & 80.9 & 68.8 & 76.5 & 77.3 & 66.9 & 54.5 & 80.9 & 73.8 & 58.7 & 84.1 & 71.5 \\
LS \cite{muller2019does} & 58.0 & 78.7 & 82.0 & 69.3 & 79.7 & 78.6 & 68.3 & 55.3 & 81.9 & 73.8 & 59.7 & 83.9 & 72.4 \\
NLNL \cite{kim2019nlnl} & 56.4 & 77.5 & 80.1 & 67.4 & 75.6 & 76.3 & 65.5 & 54.8 & 80.5 & 73.4 & 59.0 & 83.5 & 70.8 \\
SCE \cite{wang2019symmetric} & 54.5 & 78.3 & 80.6 & 63.8 & 76.7 & 76.2 & 65.3 & 50.4 & 80.6 & 70.5 & 53.7 & 82.8 & 69.4 \\
APL \cite{ma2020normalized} & 57.2 & 77.8 & 82.1 & 68.9 & 76.8 & 78.5 & 67.7 & 54.1 & 81.5 & 73.1 & 58.4 & 84.1 & 71.7 \\
LOW \cite{santiago2021low} & 58.1 & 78.8 & 82.1 & 69.5 & 79.7 & 78.5 & 68.2 & 55.3 & 81.9 & 73.8 & 59.5 & 83.9 & 72.4 \\
PolyLoss \cite{lengpolyloss} & 57.9 & 78.6 & 82.0 & 69.7 & 79.7 & 78.5 & 67.9 & 55.0 & 81.9 & 73.5 & 59.6 & 83.9 & 72.3 \\
ANL \cite{ye2023active} & 57.1 & 77.6 & 80.3 & 67.5 & 75.6 & 76.4 & 65.7 & 54.9 & 80.6 & 73.4 & 59.2 & 83.8 & 71.0 \\
AUL \cite{zhou2023asymmetric} & 56.0 & 77.9 & 81.6 & 66.6 & 79.0 & 77.7 & 66.0 & 52.4 & 81.1 & 72.6 & 57.5 & 84.0 & 71.0 \\
CE$_\epsilon$+MAE \cite{wang2024epsilon} & 56.9 & 77.9 & 80.6 & 68.4 & 76.0 & 76.9 & 66.4 & 54.4 & 80.8 & 73.6 & 59.2 & 84.0 & 71.3 \\
CE+OGC \cite{ye2025optimized} & 58.6 & 79.0 & 82.1 & 69.9 & 79.1 & 78.6 & 68.6 & 55.0 & 81.6 & 73.8 & 60.0 & 84.1 & 72.5 \\
\midrule
CBS & 57.9 & 78.3 & 81.7 & 68.7 & 79.6 & 78.3 & 68.3 & 55.4 & 82.1 & 72.8 & 59.4 & 84.2 & 72.2 \\
SRS & 57.4 & 78.7 & 81.6 & 68.4 & 79.5 & 78.5 & 68.0 & 56.1 & 81.9 & 73.5 & 58.4 & 84.5 & 72.2 \\
PBS & 57.6 & 78.5 & 82.2 & 68.4 & 79.7 & 78.6 & 68.4 & 56.3 & 82.2 & 73.5 & 59.5 & 83.9 & 72.4 \\
\midrule
\textbf{NS-WS (ours)} & \textbf{59.1} & \textbf{79.1} & \textbf{82.4} & \textbf{70.0} & \textbf{79.8} & \textbf{78.8} & \textbf{68.8} & \textbf{57.0} & \textbf{82.3} & \textbf{73.9} & \textbf{60.1} & \textbf{85.0} & \textbf{73.0}\\
\bottomrule
\end{tabular}
\label{tab:sfda}
\end{table*}

We adopt the same WS-4 configuration on the standard CIFAR10 and CIFAR100 datasets \cite{krizhevsky2009learning} and evaluate the proposed method using five widely-used networks, with results shown in Table \ref{tab:cifar}. The results show that across different networks on both datasets, NS-WS achieves an accuracy improvement of 0.3\% to 2.2\%, further confirming its generalization effectiveness in image classification tasks.

\subsubsection{Visual Emotion Recognition}
Given the varying degrees of class imbalance in these datasets, we adopt the NS-LF strategy to give greater attention to samples at a competitive disadvantage. We conduct experiments on six commonly used emotion recognition datasets and select six classic deep networks as baseline models to evaluate the effect of introducing the NS-LF strategy. As shown in Table \ref{tab:emotion}, different networks exhibit noticeable performance differences across datasets. NS-LF does not rely on a specific network architecture and consistently improves recognition accuracy, with average gains ranging from 0.6\% to 1.8\%, which demonstrates its effectiveness for emotion recognition.

\subsubsection{Source-Free Domain Adaptation (SFDA)}
Given the significant pseudo-label noise issue in SFDA tasks, we employ the NS-WS strategy to suppress the impact of noisy samples. In the experiments, we use SHOT \cite{liang2020we} as the baseline and compare NS-WS with existing methods across all 12 sub-tasks of Office-Home \cite{venkateswara2017deep}. As shown in Table \ref{tab:sfda}, the majority of these existing methods provide minimal or no performance improvement, with several leading to pronounced performance drops in the SFDA task. In contrast, NS-WS increases the average accuracy by 0.7\% and outperforms all competitors on every sub-task. Since SFDA settings often involve noisy pseudo-labels, these results highlight the superior effectiveness of NS-WS for network optimization in such challenging scenarios.

\subsubsection{Long-Tailed Classification}
For this typical class-imbalanced task, the NS-LF strategy is employed to enhance the model's learning capability for minority class samples. We employ ResNet-32 \cite{he2016deep} as the baseline and conduct comparative experiments on the CIFAR-LT-10 and CIFAR-LT-100 datasets \cite{cui2019class}. Each dataset includes four imbalance factor settings (200, 100, 50, 10), representing the ratio between the number of samples in the most frequent and the least frequent classes. The results in Table \ref{tab:longtail} show that as the imbalance ratio increases from 10 to 200, the accuracy of all methods exhibits a declining trend. Compared to the baseline, a few methods, such as Focal Loss and LOW, achieve improvements under some imbalance factors, though the gains are limited, and most methods show negligible or even degraded performance. In contrast, our NS-LF delivers consistent improvements across all imbalance factor settings, demonstrating its effectiveness in optimizing deep networks under long-tailed data distributions.

\begin{table}[t]
\centering
\caption{Performance comparison on the \textbf{CIFAR-LT-10/100} datasets.}
\setlength{\tabcolsep}{1.8mm}
\begin{tabular}{@{}lcccc|cccc@{}}
\toprule
\multirow{2}{*}{Method} & \multicolumn{4}{c|}{\textbf{CIFAR-LT-10}} & \multicolumn{4}{c}{\textbf{CIFAR-LT-100}}\\
\cmidrule(lr){2-5} \cmidrule(lr){6-9}
{} & 200 & 100 & 50 & 10 & 200 & 100 & 50 & 10 \\
\midrule
Baseline & 64.0 & 69.5 & 72.7 & 86.7 & 33.6 & 38.4 & 42.1 & 53.8 \\
Focal Loss \cite{lin2017focal} & 60.2 & 69.5 & 73.4 & 86.9 & 33.6 & 37.6 & 43.6 & 55.4 \\
GCE \cite{zhang2018generalized} & 46.3 & 60.3 & 63.8 & 86.2 & 18.6 & 21.8 & 25.3 & 36.8 \\
LS \cite{muller2019does} & 62.5 & 70.3 & 74.8 & 86.8 & 32.9 & 37.8 & 42.9 & 56.0 \\
NLNL \cite{kim2019nlnl} & 33.0 & 35.1 & 39.0 & 52.6 & 1.9 & 1.6 & 1.6 & 2.2 \\
SCE \cite{wang2019symmetric} & 59.3 & 66.2 & 74.4 & 10.0 & 1.0 & 31.0 & 36.3 & 53.6 \\
APL \cite{ma2020normalized} & 29.9 & 30.8 & 31.3 & 52.6 & 6.9 & 9.2 & 11.3 & 15.1 \\
LOW \cite{santiago2021low} & 63.0 & 70.3 & 74.4 & 87.0 & 33.2 & 38.9 & 43.1 & 56.2 \\
PolyLoss \cite{lengpolyloss} & 59.4 & 69.7 & 74.5 & 86.6 & 34.2 & 39.6 & 42.9 & 56.0 \\
ANL \cite{ye2023active} & 53.1 & 67.5 & 74.8 & 85.3 & 15.7 & 17.2 & 18.6 & 28.7 \\
AUL \cite{zhou2023asymmetric} & 28.4 & 28.4 & 35.4 & 45.4 & 3.7 & 3.6 & 5.2 & 4.5 \\
CE$_\epsilon$+MAE \cite{wang2024epsilon} & 53.4 & 61.4 & 65.2 & 86.1 & 26.0 & 30.2 & 34.9 & 48.8 \\
CE+OGC \cite{ye2025optimized} & 42.0 & 43.9 & 44.2 & 77.1 & 24.3 & 27.2 & 36.3 & 45.9 \\
\midrule
CBS & 61.0 & 68.5 & 75.1 & 86.6 & 26.8 & 30.5 & 38.5 & 53.9 \\
SRS & 61.6 & 69.9 & 75.2 & 86.5 & 32.2 & 37.4 & 41.6 & 55.1 \\
PBS & 63.6 & 68.7 & 75.2 & 86.8 & 28.7 & 31.4 & 38.3 & 54.6 \\
\midrule
\textbf{NS-LF (ours)} & \textbf{65.9} & \textbf{71.2} & \textbf{75.3} & \textbf{87.3} & \textbf{35.4} & \textbf{39.7} & \textbf{43.9} & \textbf{56.4} \\
\bottomrule
\end{tabular}
\label{tab:longtail}
\end{table}

\begin{table}[t]
\centering
\caption{Evaluation of five different grouped stitching configurations.}
\begin{tabular}{cccccc}
\toprule
\textbf{Group} & \textbf{1$\times$2} & \textbf{2$\times$2} & \textbf{2$\times$4} & \textbf{4$\times$2} & \textbf{4$\times$4} \\
\midrule
Acc (\%) & 72.0 & \textbf{72.1} & 71.7 & 71.4 & 71.7 \\ 
\bottomrule
\end{tabular}
\label{tab:group}
\end{table}

\subsection{Ablation and Analysis}

\subsubsection{Ablation on Grouping Strategy and Hyperparameters}
We conduct an experimental analysis on the CIFAR-100 dataset using ResNet-110, focusing on the image stitching grouping strategies as well as parameters $\sigma$ and $\rho$ involved in the proposed method.

(a) We evaluate five different grouped stitching configurations, including 1$\times$2 (2 images), 2$\times$2 (4 images), 2$\times$4 (8 images), 4$\times$2 (8 images), and 4$\times$4 (16 images) layouts. As shown in Table \ref{tab:group}, our evaluation reveals that group sizes of 2 and 4 both achieve better and comparable performance. Therefore, we adopt the 2/4-group configuration across all experiments.

(b) We evaluate the base value $\sigma$ in Eq. (\ref{eq:weight}). As shown in Table \ref{tab:sigma}, setting $\sigma=0$ leads to significant performance degradation, primarily because the original NS scores of some samples are too small (close to zero), which hinders effective network optimization. The model achieves optimal performance when $\sigma$ is set to 0.8. Given that our study involves multiple tasks with substantial differences in dataset distributions, we recommend adopting a similar hyperparameter selection strategy to determine the optimal $\sigma$ value for different applications.

\begin{table}[htbp]
\centering
\caption{Performance with different values of $\sigma$.}
\begin{tabular}{cccccccc}
\toprule
\textbf{$\sigma$} & \textbf{0.0} & \textbf{0.1} & \textbf{0.5} & \textbf{0.8} & \textbf{1.0} & 1.5 & 1.8 \\
\midrule
Acc (\%) & 67.5 & 70.9 & 71.3 & \textbf{72.1} & 71.7 & 71.5 & 71.4 \\ 
\bottomrule
\end{tabular}
\label{tab:sigma}
\end{table}

(c) On the CIFAR-100 dataset, we adopt the NS-based winner-strengthening strategy with the parameter $\rho$ set to 1. To validate the rationality of this parameter choice, we further evaluate the model performance with different values of $\rho$. As shown in Table \ref{tab:rho}, the model achieves optimal performance when $\rho=1$. Based on this empirical result, we set $\rho$ at 1 in our experiments without further fine-tuning of this parameter.

\begin{table}[htbp]
\centering
\caption{Performance with different values of $\rho$.}
\begin{tabular}{cccccccc}
\toprule
\textbf{$\rho$} & \textbf{0.0} & \textbf{0.1} & \textbf{0.5} & \textbf{0.8} & \textbf{1.0} & 1.5 & 1.8 \\
\midrule
Acc (\%) & 70.3 & 71.3 & 71.6 & 71.8 & \textbf{72.1} & 71.9 & 71.8 \\ 
\bottomrule
\end{tabular}
\label{tab:rho}
\end{table}

\subsubsection{Statistical Stability}
In our experiments, we perform multiple runs using three random seeds (2024, 2025, 2026) and report the average performance. We systematically compile the mean results and standard deviations across the 12 public datasets considered in this study, as shown in Table \ref{tab:stat}. Overall, the performance gains achieved by our method are significantly greater than random fluctuations, indicating that the observed improvements are statistically significant.

\begin{table}[htbp]
\centering
\caption{Mean and standard deviation of the performance on the 12 public datasets.}
\begin{tabular}{lclc}
\toprule
\textbf{Dataset} & \textbf{Mean$\pm$Std} & \textbf{Dataset} & \textbf{Mean$\pm$Std} \\
\midrule
ImageNet-1K   & 73.9$\pm$0.05 & Instagram     & 83.7$\pm$0.21 \\
CIFAR-10      & 91.8$\pm$0.08 & FI            & 64.2$\pm$0.26 \\
CIFAR-100     & 70.9$\pm$0.22 & EmoSet        & 75.1$\pm$0.12 \\
Twitter I     & 77.0$\pm$0.45 & Office-Home   & 73.0$\pm$0.19 \\
Twitter II    & 73.3$\pm$0.34 & CIFAR-LT-10   & 74.9$\pm$0.43 \\
Flickr        & 85.5$\pm$0.11 & CIFAR-LT-100  & 43.9$\pm$0.18 \\
\bottomrule
\end{tabular}
\label{tab:stat}
\end{table}

\subsubsection{Additional Evaluation on Hard Samples}
To examine whether the NS-WS strategy adversely affects the handling of hard samples, we conduct a dedicated evaluation on the ImageNet-Hard dataset \cite{taesiri2023imagenet}. This dataset comprises numerous images that are generally challenging for existing models to classify correctly, making it suitable for evaluating the robustness of methods on hard samples. Using ResNet-18 as the baseline, we perform experiments as summarized in Table \ref{tab:hard}. It can be observed that after incorporating the NS-WS strategy, the model's performance on hard samples do not exhibit significant degradation. This indicates that the proposed method poses a low risk of marginalizing legitimate hard samples or minority patterns.

\begin{table}[htbp]
\centering
\caption{Results on the \textbf{ImageNet-Hard} dataset.}
\begin{tabular}{lcc}
\toprule
\textbf{Dataset} & \textbf{ResNet-18} & \textbf{w/ NS-WS} \\
\midrule
ImageNet-Hard & 9.9 & \textbf{10.2} \\ 
\bottomrule
\end{tabular}
\label{tab:hard}
\end{table}

\begin{figure*}[t]
\centering
\footnotesize
    \begin{tabular}{ccc}
    \hspace{-12pt}
    \subfloat[CIFAR-10]{
    \includegraphics[width=0.45\textwidth]{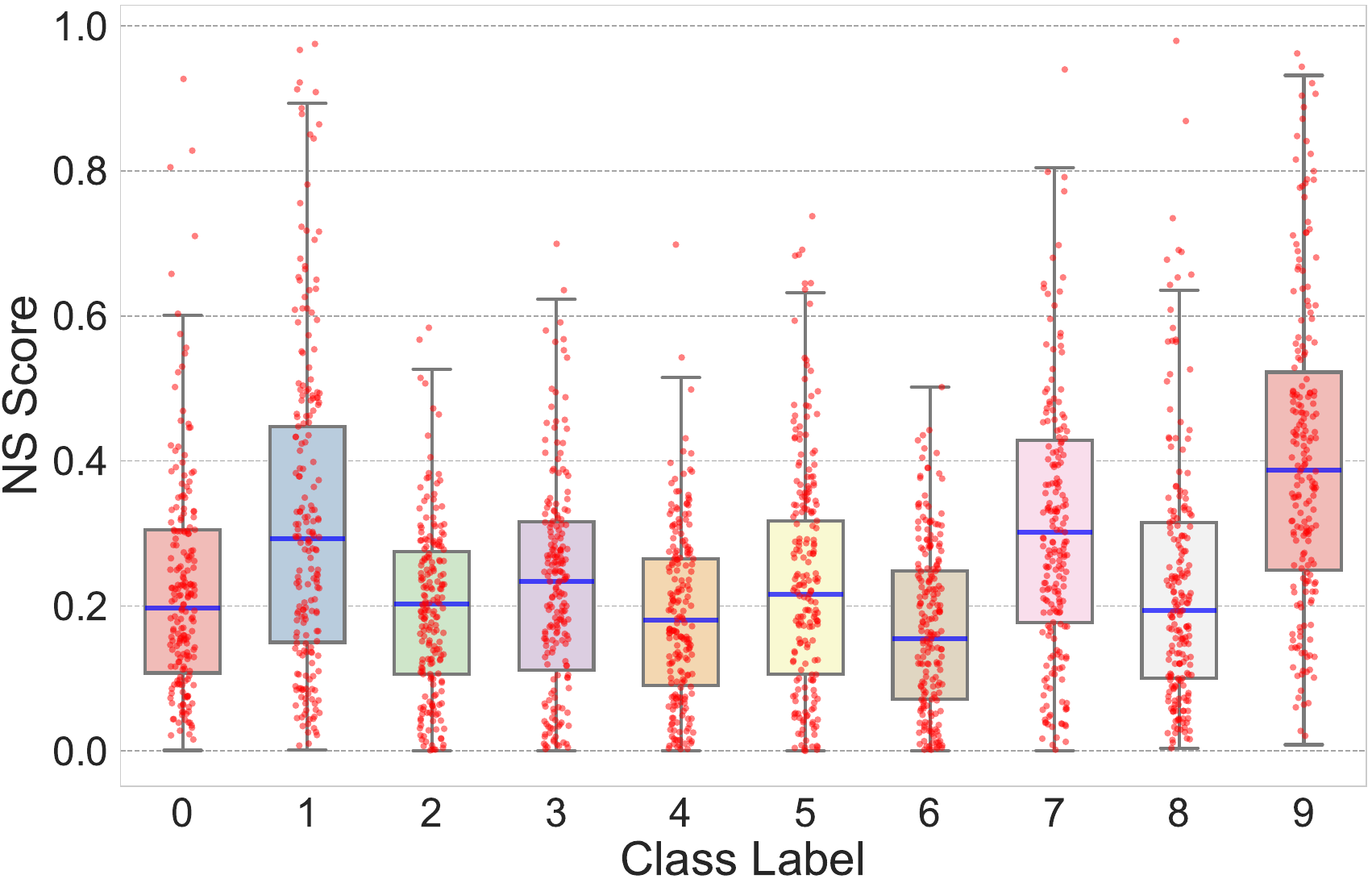}
    } &
    \subfloat[FI]{
    \includegraphics[width=0.45\textwidth]{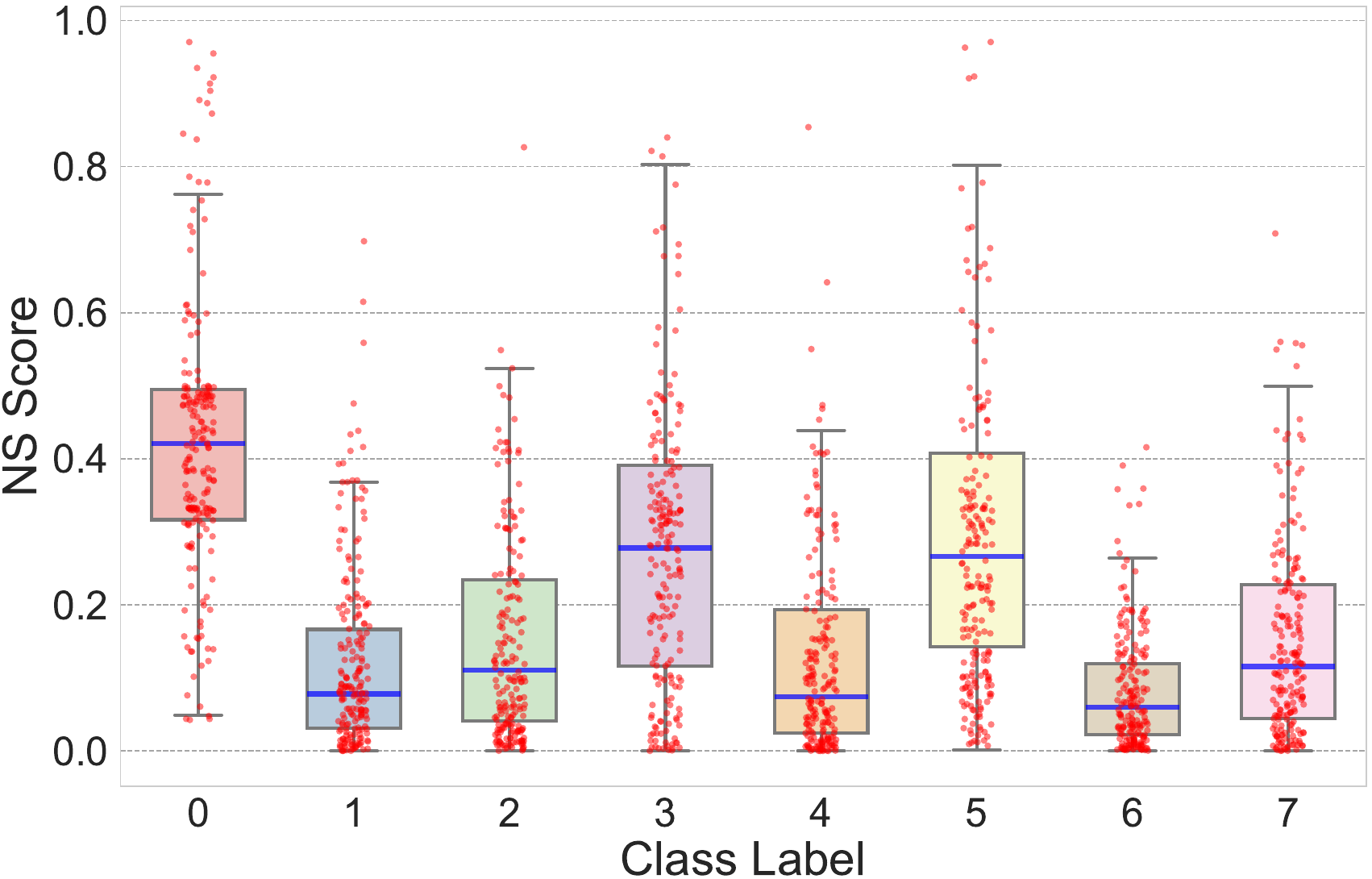}
    }
    \end{tabular}
\caption{Visualization of NS score distributions by category on the CIFAR-10 and FI datasets. Red points denote training samples, and the blue solid line within each box indicates the median NS score of all samples in that class.}
\label{fig:distribution}
\end{figure*}

\subsubsection{Extension to Text Classification}
We conduct additional experiments on text classification tasks, using two public datasets (IMDB \cite{maas2011learning} and AG News \cite{zhang2015character}) and employing TextCNN \cite{kim2014convolutional} and BERT-base \cite{devlin2019bert} as baseline models for evaluation. Specifically, for text-based samples, we concatenate four text instances into a single input, which is then converted into sequence or token representations. Subsequently, we compute the NS score for each text via model inference and apply the NS-LF strategy to adjust the training loss accordingly. The experimental results presented in the Table \ref{tab:text} demonstrate that our method achieves consistent performance improvements in text classification tasks, indicating its cross-modal applicability.

\begin{table}[htbp]
\centering
\caption{Results on the \textbf{IMDB} and \textbf{AG News} datasets.}
\begin{tabular}{lcccc}
\toprule
\textbf{Dataset} & \textbf{TextCNN} & \textbf{w/ NS-LF} & \textbf{BERT-base} & \textbf{w/ NS-LF} \\
\midrule
IMDB & 75.6 & \textbf{76.6} & 91.8 & \textbf{92.2} \\
AG News & 90.3 & \textbf{91.0} & 94.4 & \textbf{94.7} \\ 
\bottomrule
\end{tabular}
\label{tab:text}
\end{table}

\begin{table}[t]
\centering
\caption{Computational overhead of NS in terms of GPU memory usage and per-epoch training time.}
\setlength{\tabcolsep}{1.2mm}
\begin{tabular}{@{}lcccc@{}}
\toprule
\multirow{2}{*}{Method} & \multicolumn{2}{c}{\textbf{CIFAR-10}} & \multicolumn{2}{c}{\textbf{CIFAR-100}} \\
\cmidrule(lr){2-3} \cmidrule(lr){4-5}
& Memory & Time & Memory & Time \\
& (MB) & (s) & (MB) & (s) \\
\midrule
ResNet-110 & 1,564 & 23.81 & 1,564 & 23.93 \\
w/ NS & 1,588 {\scriptsize(\textcolor[rgb]{0.2,0.6,0.4}{$\uparrow$1.5\%})}
      & 26.24 {\scriptsize(\textcolor[rgb]{0.2,0.6,0.4}{$\uparrow$10.2\%})}
      & 1,588 {\scriptsize(\textcolor[rgb]{0.2,0.6,0.4}{$\uparrow$1.5\%})}
      & 26.97 {\scriptsize(\textcolor[rgb]{0.2,0.6,0.4}{$\uparrow$12.7\%})} \\
\bottomrule
\end{tabular}
\label{tab:computation}
\end{table}

\begin{table}[t]
\centering
\caption{Per-step runtime breakdown of NS.}
\setlength{\tabcolsep}{5pt}
\begin{tabular}{@{}lcc@{}}
\toprule
\textbf{Operation} & \textbf{CIFAR-10} & \textbf{CIFAR-100} \\
\midrule
Stitching \& scaling & 0.14 s & 0.14 s \\
Model inference      & 2.31 s & 2.87 s \\
\bottomrule
\end{tabular}
\label{tab:cost_detail}
\end{table}

\subsubsection{Computational Overhead}
We evaluated the computational overhead introduced by the proposed NS method on the CIFAR-10 and CIFAR-100 datasets using the ResNet-110 network, with a primary focus on GPU memory usage and training time. The results are shown in Table \ref{tab:computation}. In terms of memory consumption, although the number of classes differs between the two datasets, leading to a variation in the number of parameters in the final layer, this difference is negligible due to the minimal proportion of these parameters in the overall model. Experiments show that the baseline memory usage remains consistent across both datasets. On this basis, NS introduces only a 1.5\% additional memory overhead, demonstrating high memory efficiency. Regarding training time, we report the average results over 10 independent runs. Compared to the baseline, the introduction of NS increases training time by 10.2\% and 12.7\% on CIFAR-10 and CIFAR-100, respectively. This overhead comes mainly from the image stitching, scaling, and model inference in the NS module. The timing analysis for each step in the NS computation process is presented in Table \ref{tab:cost_detail}. The results demonstrate that the additional computational cost primarily stems from the model inference stage, while the time overhead introduced by image stitching and scaling operations is negligible. Still, the increase remains within an acceptable range. It does not form a significant computational bottleneck.

In summary, NS is a plug-and-play training module that delivers consistent performance gains with negligible memory overhead and only modest additional training cost, highlighting its practical applicability.

\subsubsection{NS Score Distribution}
On the CIFAR-10 dataset, we employ the ResNet-110 model, while on the FI dataset, we utilize the ResNet-50 model to statistically analyze the distribution of NS scores across categories (taking one training epoch as an example), as shown in Fig. \ref{fig:distribution}. To illustrate competitive divergence among samples more clearly, we randomly select 200 samples from each category for visualization. The box plots in Figs. \ref{fig:distribution} (a) and (b) reveal marked differences in NS score distributions across categories, reflecting competitive relationships between classes. Meanwhile, the scatter plots of samples within each category reveal notable variation in NS scores, even among samples from the same class, demonstrating intra-class competition. These results suggest that competitive differences among samples, as measured by NS scores, effectively mirror the interspecific and intraspecific competition in ecosystems. Consequently, the NS score serves as an effective metric for quantifying sample competitiveness and establishes a basis for incorporating an evolution-inspired dynamic equilibrium into the network optimization process.

\begin{figure*}[t]
\centering
\footnotesize
    \begin{tabular}{ccc}
    \hspace{-12pt}
    \subfloat[CIFAR-LT-10]{
    \includegraphics[width=0.33\textwidth]{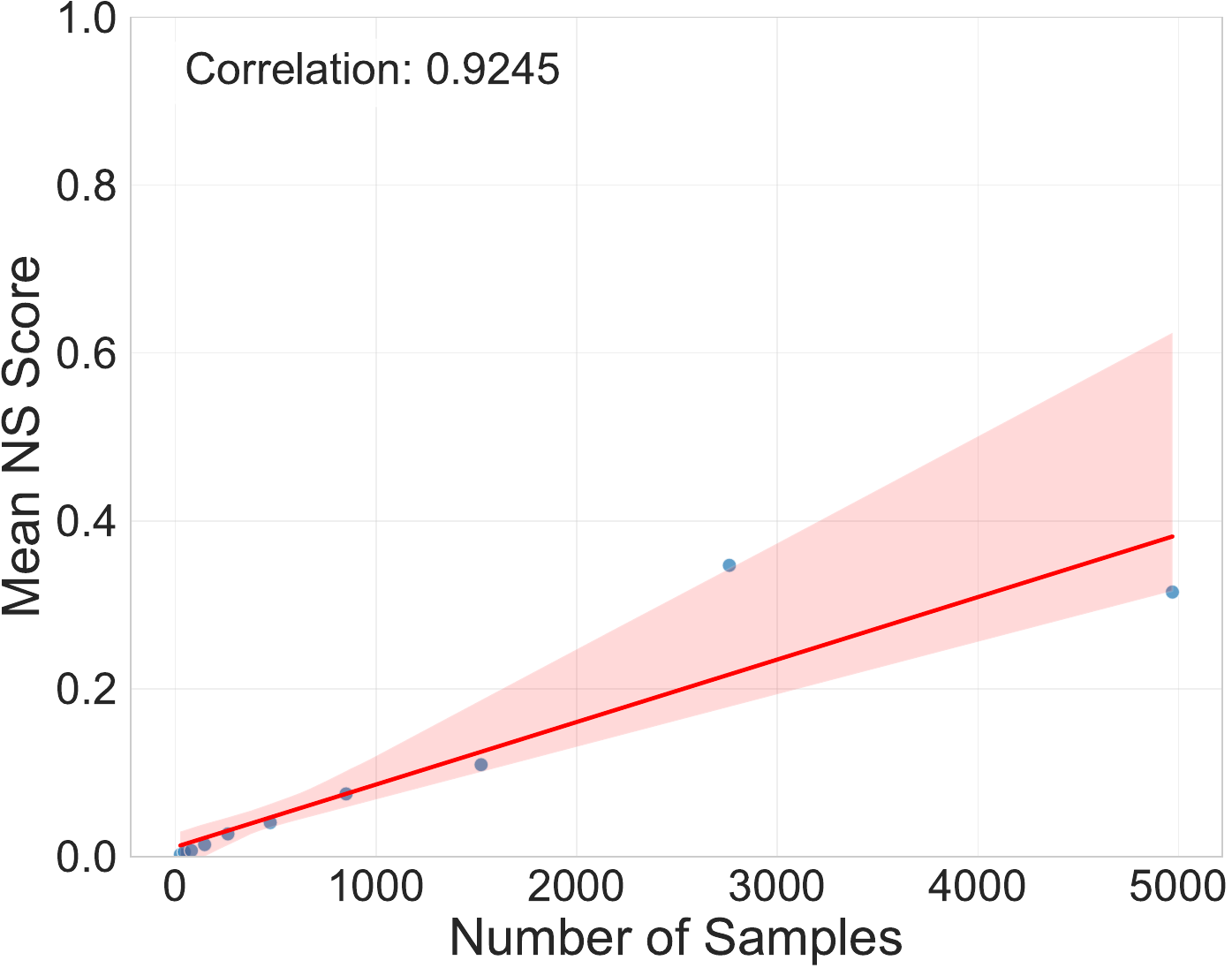}
    } &
    \hspace{-15pt}
    \subfloat[CIFAR-LT-100]{
    \includegraphics[width=0.33\textwidth]{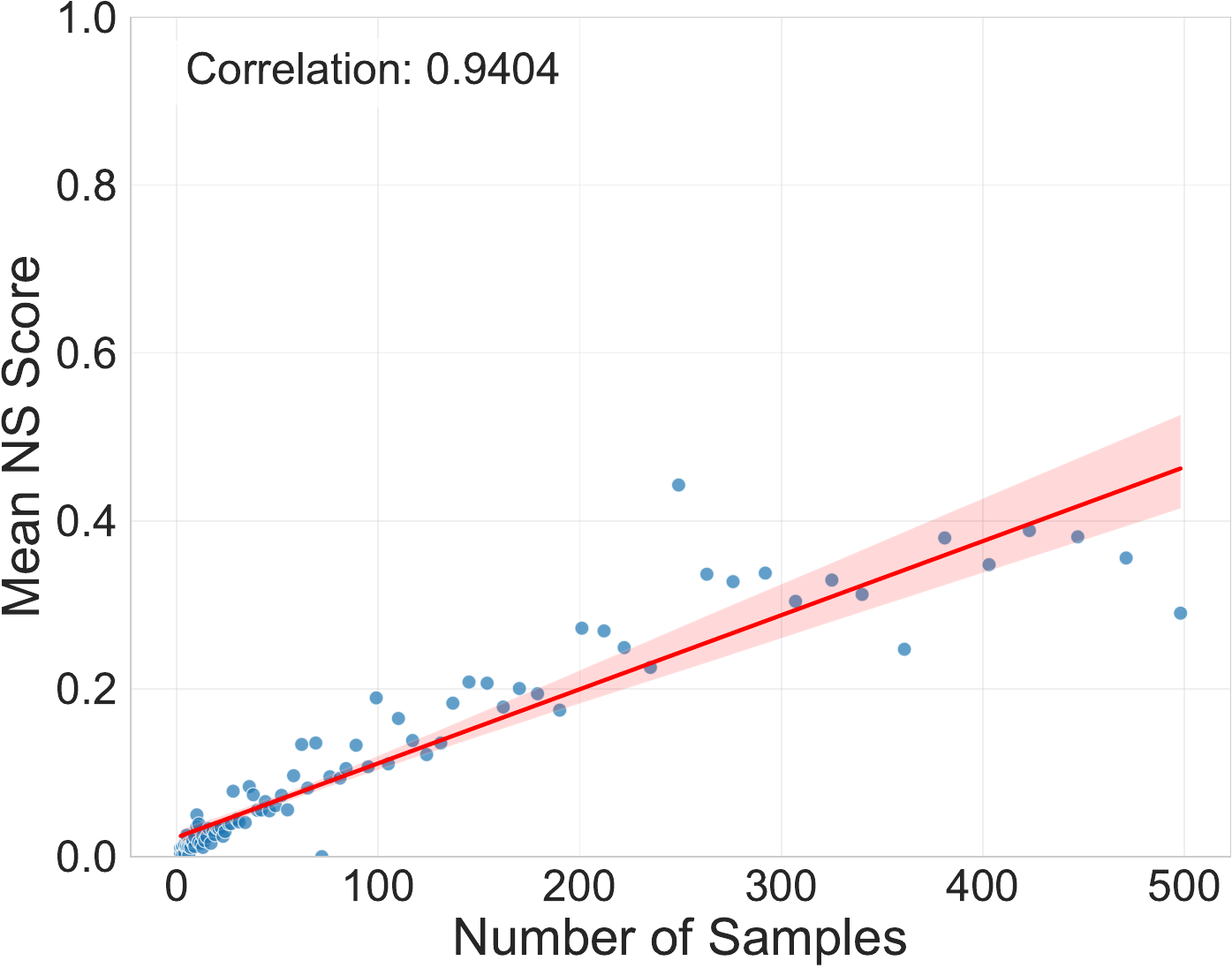}
    } &
    \hspace{-15pt}
    \subfloat[Office-Home (Cl$\rightarrow$Pr)]{
    \includegraphics[width=0.33\textwidth]{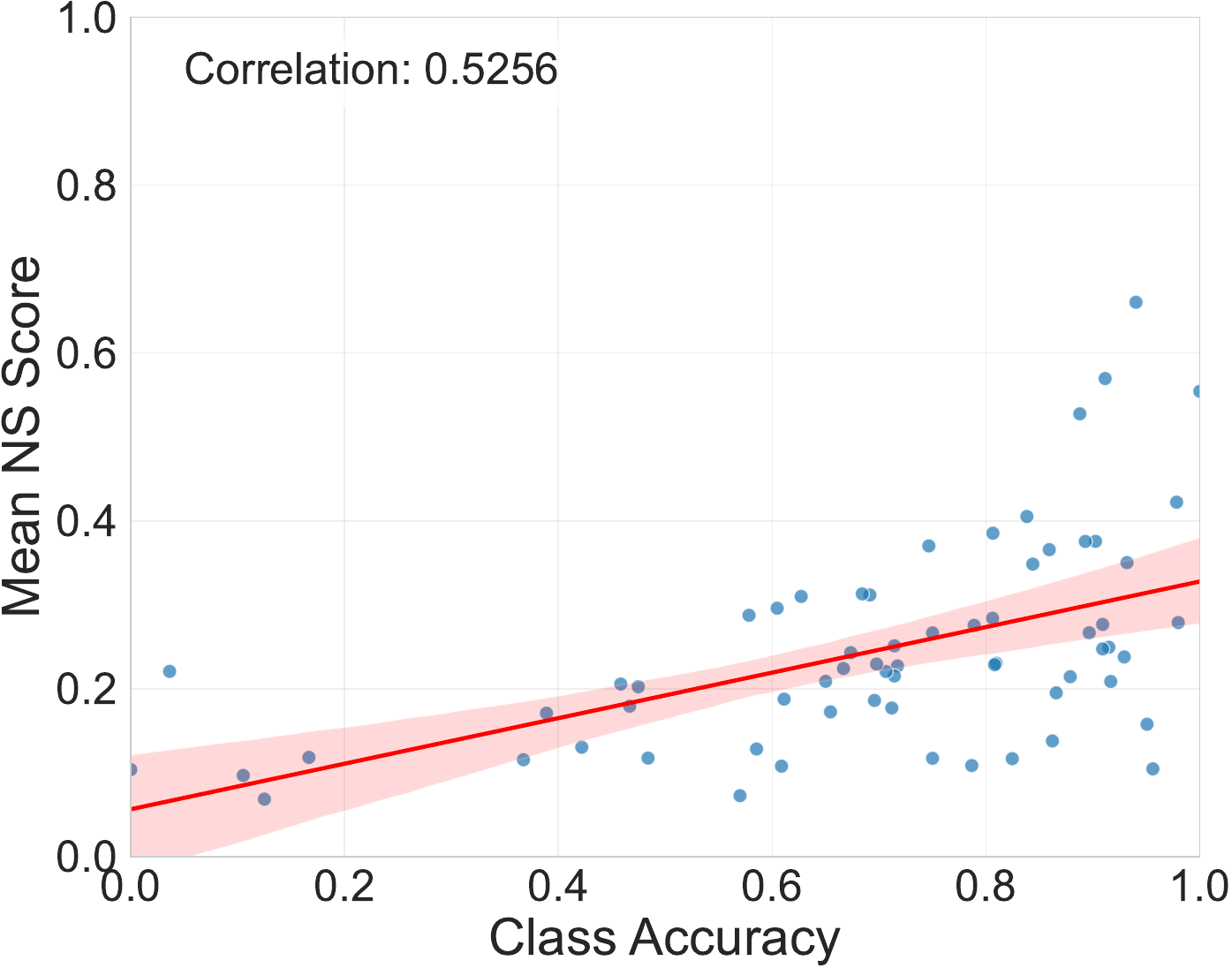}
    }
    \end{tabular}
\caption{Correlation analysis on the CIFAR-LT-10/100 and Office-Home datasets. Blue points represent the individual categories, while the red line depicts the fitted linear regression.}
\label{fig:correlation}
\end{figure*}

\subsubsection{Correlation Analysis}
We conduct class-wise analyses on the CIFAR-LT-10/100 datasets under class-imbalance settings. This allows us to examine the correlation between class sample size and mean NS score. On the Office-Home dataset under the SFDA setting, we examine the correlation between class accuracy and mean NS score. As shown in Figs. \ref{fig:correlation} (a) and (b), the mean NS score exhibits a significant positive correlation with class sample size. This indicates that classes with more samples tend to have higher average NS scores and thus a relative competitive advantage. Furthermore, Fig. \ref{fig:correlation} (c) shows a positive correlation between the mean NS score and class accuracy. In the SFDA scenario, this typically means that classes with higher pseudo-label quality (lower noise) also have higher NS scores. These observations provide empirical support for NS score-guided optimization: moderately enhancing competitively disadvantaged samples in class-imbalanced environments, and prioritizing competitively advantaged samples to provide relatively reliable supervision in SFDA. We emphasize that these findings are based on correlational rather than causal inference, and the effectiveness of our method has been validated through the aforementioned experiments.

\subsection{Limitations and Future Work}
\label{app:future}
Although the proposed NS method demonstrates consistent effectiveness across multiple tasks, the current evaluation remains focused mainly on computer vision, with only preliminary validation on text classification. As a result, the robustness and generality of NS across a broader range of learning settings have not yet been comprehensively established.

As a natural next step, it would be valuable to extend the evaluation of NS to a wider spectrum of vision problems, such as fine-grained visual classification \cite{zhang2025leaf}, few-shot learning \cite{zhang2022well}, video-based action recognition \cite{zheng2018distinctive,lu2024mixed}, and vision-based robotic manipulation \cite{zheng2024survey}. These tasks involve different forms of intra-class variation and temporal dependency, and may therefore provide a more comprehensive understanding of the strengths and limitations of NS. Beyond vision, while our current study offers only preliminary evidence in text classification, the underlying idea of competition-aware optimization may also be relevant to broader multimodal and cross-modal learning scenarios. In particular, it may provide a useful perspective for the design of optimization strategies in multimodal large language models and related systems. We leave these directions for future work.

\section{Conclusion}
Traditional deep network training methods typically apply uniform selective pressure to each sample, ignoring inherent competitive differences among them. Inspired by biological mechanisms, this paper proposes a natural selection method that explicitly simulates competitive interactions between samples for ecologically balanced optimization. Specifically, we stitch a group of images for joint prediction, evaluating the competitive strength of each sample to compute a corresponding natural selection score. This score then dynamically modulates each sample's training loss, thereby enhancing overall network optimization. Experimental results across 12 public datasets covering four types of computer vision tasks demonstrate that the proposed method achieves consistent performance gains without relying on specific network architectures. We emphasize that the NS method offers notable usability and generalization potential. Its applicability is not limited to the tasks validated in this paper and shows promise for extension to a broader range of research areas.

\bibliographystyle{IEEEtran}
\bibliography{refs}

\vfill

\end{document}